\documentclass{article}

\usepackage[preprint]{neurips_2026}
\usepackage{amsmath}

\usepackage[utf8]{inputenc} 
\usepackage[T1]{fontenc}    
\usepackage{hyperref}       
\usepackage{url}            
\usepackage{booktabs}       
\usepackage{amsfonts}       
\usepackage{nicefrac}       
\usepackage{microtype}      
\usepackage{xcolor}         

\usepackage{graphicx}
\usepackage{tikz}
\usetikzlibrary{positioning, arrows.meta, shapes.geometric, calc}
\usepackage{arydshln}

\PassOptionsToPackage{numbers, compress}{natbib}
\usepackage{multirow}

\title{HH-SAE: Discovering and Steering Hierarchical Knowledge of Complex Manifolds}

\author{%
  Honghan Wu\\
  University of Glasgow, Glasgow, UK \\
  \texttt{honghan.wu@glasgow.ac.uk} \\
  https://knowlab.github.io/ 
  \And
  Tianyan Wang \\
  University of International Relations, Beijing, China \\
  \texttt{tianyanwang1@gmail.com} \\
  https://knowlab.github.io/ 
  \AND
  Jiacong Mi \\
  University of Glasgow, Glasgow, UK \\
  \texttt{j.mi.1@research.gla.ac.uk} \\
  https://knowlab.github.io/ 
  \AND
  Zhoyang Jiang \\
  University of Glasgow, Glasgow, UK \\
  \texttt{3167645J@student.gla.ac.uk} \\
  https://knowlab.github.io/ 
  \And
  Yunsoo Kim \\
  University College London, London, UK \\
  \texttt{yunsoo.kim.23@ucl.ac.uk} \\
  https://knowlab.github.io/ 
}

\begin{document}

\maketitle

\begin{abstract}
Rare semantic innovations in high-dimensional, mission-critical domains are often obscured by dense background contexts, a challenge we define as \textit{feature density conflict}. We introduce the \textbf{Hybrid Hierarchical SAE (HH-SAE)} to resolve this by factorizing manifolds into a nested hierarchy of \textbf{Contextual} ($L_0$), \textbf{Atomic} ($f_1$), and \textbf{Compository} ($f_2$) tiers. Evaluating across disparate manifolds, HH-SAE demonstrates superior resolution by \textbf{``fracturing'' administrative clinical labels into physiological modes} and achieving a peak \textbf{cross-domain zero-shot AUC of 0.9156 in fraud detection}. Path ablation confirms the architecture's structural necessity, revealing a 13.46\% utility collapse when contextual subtraction is removed. Finally, knowledge-steered synthesis achieves a +9.9\% AUPRC lift over state-of-the-art generators, proving that HH-SAE effectively prioritizes high-order mechanistic innovation over environmental proxies to enable high-precision discovery in high-stakes environments.
\end{abstract}

\section{Introduction}

High-dimensional data in mission-critical systems—ranging from clinical trajectories to financial transactional manifolds—rarely exists as a flat distribution. Instead, it is a nested composition of two distinct regimes: a high-density \textbf{Contextual} background (steady-state populations or demographic ``shadows'') and a sparse, low-density \textbf{Innovation} manifold (rare, synergistic signals that drive system shifts). Standard dictionary learning methods~\cite{mairal2009online}, such as Sparse Autoencoders (SAEs)~\cite{bricken2023towards}, often conflate these regimes. This leads to a \textbf{feature density conflict}, where the model's capacity is consumed by redundant background variables, effectively smoothing over the rare, synergistic interactions that constitute true domain knowledge~\cite{bricken2023towards,openai2024scaling}.

In this work, we propose the \textbf{Triplet Knowledge Hypothesis}: that complex manifolds are factorizable into three distinct semantic tiers. First, a high-density \textbf{Contextual} manifold ($L_0$) representing the environmental baseline. Second, an \textbf{Atomic} manifold ($f_1$) of sparse, discrete innovations. Finally, a \textbf{Compository} manifold ($f_2$) where these atoms aggregate into synergistic motifs or ``grammars'' that drive critical system shifts. We introduce the \textbf{HH-SAE} (Hybrid Hierarchical Sparse Autoencoder), a dual-path architecture that resolves density conflicts by using a low-rank dense bottleneck to ``stiffen'' and subtract the contextual background, allowing a tiered sparse hierarchy to act as a \textit{Sparse Miner} for hierarchical innovations.

Beyond discovery, this architecture enables a unique capacity for \textbf{Mechanistic Knowledge Synthesis}. Unlike black-box generative models that interpolate across an opaque latent space, the HH-SAE facilitates \textbf{Knowledge-Steering}: by perturbing specific hierarchical nodes—such as clinical ``atomic knobs'' or financial ``compository motifs''—researchers can generate high-fidelity synthetic data that adheres to the discovered structural logic of the domain. This steering capability serves as a rigorous litmus test for the model's understanding; if the hierarchy captures the system's underlying mechanistic links rather than mere statistical correlation, the resulting synthetic samples must preserve the complex, synergistic trajectories of the original manifold.

Our key contributions are:
\begin{itemize}
    \item \textbf{The Triplet Knowledge Hypothesis:} We propose a novel discovery regime that factorizes high-dimensional manifolds into a nested hierarchy of \textbf{Contextual} ($L_0$), \textbf{Atomic} ($f_1$), and \textbf{Compository} ($f_2$) tiers, effectively resolving the feature density conflict that consumes the capacity of standard SAEs.
    \item \textbf{Structural Resolution of Density Conflict:} We introduce a dual-path hybrid architecture that disentangles dense contextual backgrounds from sparse semantic innovations. Path ablation confirms this design as a structural requirement, revealing a \textbf{13.46\%} utility collapse when contextual subtraction is removed.
    \item \textbf{Cross-Domain Zero-Shot Resilience:} Validation across disparate manifolds---MIMIC-IV clinical generalization and IEEE-CIS financial zero-shot transfer---shows significant predictive gains (up to $+0.043$ AUC), proving the model's ability to maintain mechanistic fidelity across distribution shifts.
    \item \textbf{Mechanistic Knowledge-Steered Synthesis:} We provide a methodology for utilizing discovered hierarchical motifs to steer data synthesis, achieving a \textbf{+9.9\%} relative improvement in AUPRC. This validates that HH-SAE captures the underlying structural logic of a domain rather than mere statistical correlations.
\end{itemize}

\section{Methods}
\label{sect:method}
\begin{figure}[h]               
    \centering                  
    \includegraphics[width=0.99\textwidth]{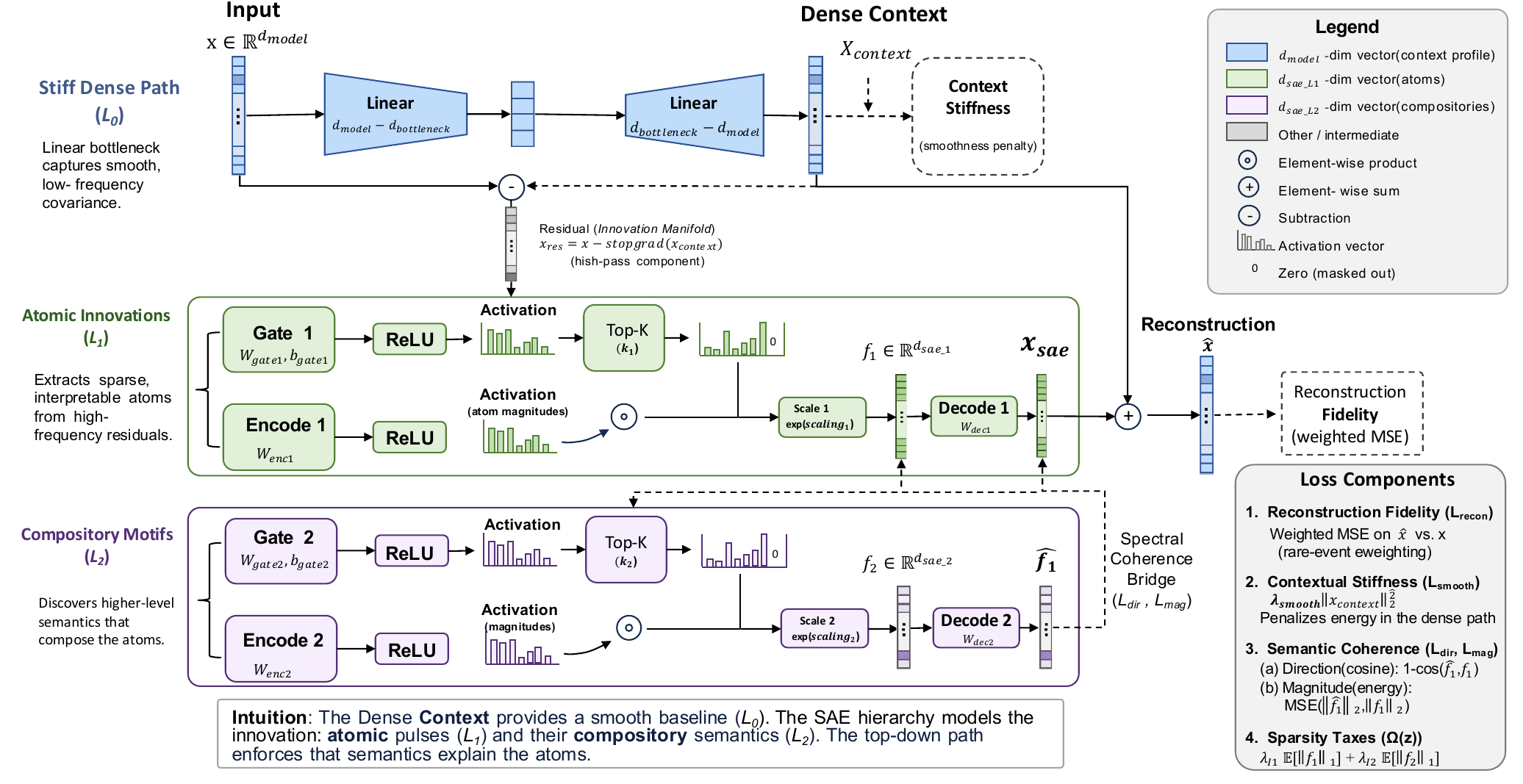} 
    \caption{\textbf{HH-SAE Architectural and Objective Schematic.} The framework factorizes the manifold into a functional triplet: a \textbf{Stiff Dense Path} ($L_0$) captures contextual backgrounds under stiffness constraints ($\mathcal{L}_{smooth}$), while a tiered sparse hierarchy isolates \textbf{Atomic Innovations} ($L_1$) and synthesizes \textbf{Compository Motifs} ($L_2$). A \textbf{Spectral Coherence Bridge}($\mathcal{L}_{dir}, \mathcal{L}_{mag}$) ensures hierarchical grounding by forcing motifs to provide a generative explanation for atomic activations. Streams are fused via a centered summation node for hybrid fidelity reconstruction ($\mathcal{L}_{recon}$).}
    \label{fig:hybrid_arch_final}       
\end{figure}

\subsection{Architecture Design and Motivation}

The \textbf{Hierarchical Hybrid Sparse Autoencoder (HH-SAE)}, illustrated in Figure~\ref{fig:hybrid_arch_final}, factorizes complex manifolds into a functional triplet: \textbf{Contextual} backgrounds, \textbf{Atomic} innovations, and \textbf{Compository} motifs. High-dimensional distributions are typically characterized by a high-density \textit{background manifold}—containing population-wide variance—and a sparse \textit{innovation manifold} harboring rare, high-information events. HH-SAE resolves the ``feature density conflict'' via a dual-track architecture: a stiff dense path absorbs contextual priors, while a hierarchical sparse path disentangles fundamental basis elements at the \textbf{Atomic Concept Layer} ($L_1$) and synergistic interactions at the \textbf{Compository Layer} ($L_2$). By leveraging Gated SAE units~\cite{rajamanoharan2024improving}, the architecture maintains high reconstruction fidelity across disparate density regimes, uncovering multi-scale representations latent in flat sparsity models. Throughout this manuscript, we use $L_i$ to refer to the layer components and $f_i$ to denote the specific sparse feature activations produced by these layers.

\subsection{HH-SAE Architectural Specification}

The HH-SAE factorizes the input manifold $\mathbf{x} \in \mathbb{R}^D$ into an additive composition of contextual and innovative streams:
\begin{equation}
\hat{\mathbf{x}} = \text{Dec}_{dense}(\mathbf{z}_{dense}) + \text{Dec}_{sparse}(\mathbf{z}^{(1)}, \mathbf{z}^{(2)})
\end{equation}

\subsubsection{$L_0$: Stiff Contextual Stream}
The dense path utilizes a low-rank linear bottleneck $\mathbf{z}_{dense} \in \mathbb{R}^{d_{dense}}$ ($d_{dense} \ll D$) to absorb global, low-frequency variance. By enforcing a ``stiff'' linear projection without non-linearities, this stream acts as a low-pass filter, preventing the dense path from absorbing sharp clinical innovations. This ensures the sparse hierarchy acts strictly as a residual learner for high-information signal transients.

\subsubsection{Sparse Innovation Hierarchy}
The innovation path factorizes the residual signal $\mathbf{x}_{resid} = \mathbf{x} - \text{detach}(\hat{\mathbf{x}}_{cont})$ into two tiers of abstraction using Gated Sparse Autoencoder units.

\paragraph{($L_1$): Atomic Concepts}
This layer serves as an overcomplete feature extractor $\mathbf{z}^{(1)} \in \mathbb{R}^{d_1}$ ($d_1 \gg D$). Utilizing a Gated Top-$k$ Sparse Miner, it isolates monosemantic ``atomic'' units—primitive, independent signal components such as isolated physiological transients. Activation density is strictly constrained (e.g., $\approx 2.4\%$) to ensure feature disentanglement.

\paragraph{($L_2$): Compository Motifs}
The Compository Layer implements a second-order synergistic bottleneck $\mathbf{z}^{(2)} \in \mathbb{R}^{d_2}$ ($d_2 < d_1$). Rather than raw features, $L_2$ models the joint distribution of atomic activations via a non-linear composition function $\phi: \mathbf{z}^{(1)} \to \mathbf{z}^{(2)}$. This forces the synthesis of high-level ``motifs'' from primitive atoms, capturing the essential multi-feature interactions that define rare or anomalous sub-manifolds.

\subsection{Tiered Compositional Objective}
\label{sect:loss}
The HH-SAE is optimized via a tiered objective function designed to enforce a strict division of labor across the functional triplet. The total loss $\mathcal{L}_{total}$ regularizes each component according to its intended semantic role:
\begin{equation}
\mathcal{L}_{total} = \mathcal{L}_{recon} + \lambda_{s} \mathcal{L}_{smooth} + \alpha \mathcal{L}_{dir} + \beta \mathcal{L}_{mag} + \Omega(\mathbf{z})
\end{equation}

\paragraph{Weighted Global Fidelity ($\mathcal{L}_{recon}$)} 
To prioritize the discovery of high-information clinical innovations, we employ a sample-weighted MSE: $\mathcal{L}_{recon} = \mathbb{E}[w \cdot \|\mathbf{x} - \hat{\mathbf{x}}\|_2^2]$. For an input label $y \in \{0,1\}$, the bias weight is defined as $w = 1 + y(\omega_{rare} - 1)$. This decoupling results in an \textbf{Inverse Error Gradient}, where the model achieves superior fidelity for atypical sub-manifolds compared to the redundant population background.

\paragraph{Contextual Stiffness ($\mathcal{L}_{smooth}$)} 
To ensure the dense path remains a low-pass filter, we penalize the contextual reconstruction energy: $\mathcal{L}_{smooth} = \mathbb{E}[\|\hat{\mathbf{x}}_{cont}\|_2^2]$. This constraint prevents the stiff path from ``absorbing'' sharp clinical innovations, forcing high-frequency signals into the sparse hierarchical track.

\paragraph{Hierarchical Semantic Coherence ($\mathcal{L}_{dir}, \mathcal{L}_{mag}$)} 
We force the Compository Layer ($L_2$) to act as a generative explanation for the Atomic Layer ($L_1$) by aligning the predicted state $\hat{\mathbf{z}}^{(1)}$ with detached ground-truth activations $\text{sg}[\mathbf{z}^{(1)}]$:
\begin{itemize}
    \item \textbf{Directional Alignment:} $\mathcal{L}_{dir} = 1 - \mathbb{E}[\text{cos\_sim}(\hat{\mathbf{z}}^{(1)}, \text{sg}[\mathbf{z}^{(1)}])]$ ensures motifs identify the correct pattern of atomic concepts.
    \item \textbf{Magnitude Alignment:} $\mathcal{L}_{mag} = \mathbb{E}[ (\|\hat{\mathbf{z}}^{(1)}\|_2 - \|\text{sg}[\mathbf{z}^{(1)}]\|_2)^2 ]$ prevents energy discrepancy and gradient vanishing within the hierarchy.
\end{itemize}

\paragraph{Sparsity Taxes ($\Omega$)} 
To enforce a parsimonious latent grammar, we apply $\ell_1$ penalties to both sparse layers: $\Omega(\mathbf{z}) = \lambda_1 \|\mathbf{z}^{(1)}\|_1 + \lambda_2 \|\mathbf{z}^{(2)}\|_1$. This incentivizes the model to explain complex trajectories through minimal synergistic compositions, surfacing ``invisible'' interactions typically lost in unregularized spaces.

\subsection{Hierarchical Knowledge Discovery and Manifold Interpretation}
\label{sect:knowledge_discovery}

The HH-SAE functions as a \textit{Sparse Miner}, enabling the extraction of nested semantic structures latent within complex manifolds. By neutralizing dense contextual backgrounds ($L_0$), the architecture resolves the feature density conflict to isolate sparse innovations. The discovery process is formalized through a two-stage interpretive pipeline:

\paragraph{Hierarchical Semantic Projection} 
To interpret high-level motifs, we map activations from the Compository Layer ($L_2$) back to the Atomic Layer ($L_1$), decoding the ``mechanistic grammar'' of the domain. For any compository neuron $j \in \{1, \dots, d_2\}$, we identify its semantic foundation by analyzing the weight matrix $\mathbf{W}^{(2)}$. By isolating the set of atomic concepts $\{\mathbf{z}_i^{(1)}\}$ with the highest associative weights, we construct a \textit{Compository Semantic Profile}. This allows for the resolution of synergistic innovations---such as concurrent feature spikes that are individually sub-threshold but collectively significant---which would be obscured by the global context in non-hierarchical models.

\paragraph{Co-firing Affinity and Concept Modules} 
Rather than treating neurons as isolated detectors, we identify higher-order structures via a \textit{Modular Affinity Regime}. We construct an activation affinity matrix $\mathbf{A} \in \mathbb{R}^{d_2 \times d_2}$, where $\mathbf{A}_{jk}$ represents the frequency with which compository neurons $j$ and $k$ fire simultaneously across a target innovation cohort. By applying community detection algorithms to $\mathbf{A}$, we identify \textbf{Concept Modules}: coherent ensembles of $L_2$ neurons that collectively define a recurring state or motif. 

This modularity ``fractures'' monolithic contexts into high-resolution modes, allowing $L_2$ modules to resolve ``marginal'' events via synergistic co-firing rather than isolated outliers. These discovered modules serve as the fundamental units for the steered synthesis detailed in Section~\ref{sect:knowledge_synthesis}.

\subsection{Knowledge-Steered Synthesis for Mechanistic Data Generation}
\label{sect:knowledge_synthesis}
Leveraging the hierarchical structures uncovered in Section~\ref{sect:knowledge_discovery}, the HH-SAE enables \textbf{Modular Ensemble Synthesis}. Unlike standard generative models that perform unconstrained distribution mimicry, this framework generates new samples by activating coherent phenotypic clusters (Concept Modules). The process is formalized through the following assembly:

\paragraph{Contextual Carrier Initialization} 
Generation begins with the creation of diverse ``carrier'' bodies that represent the stable background manifold. We sample noise $\boldsymbol{\epsilon} \sim \mathcal{N}(0, \mathbf{I})$ and project it into the feature space using population-level statistics $(\mu_{pop}, \sigma_{pop})$. These inputs are passed through the dense path to extract a stable contextual baseline:
\begin{equation}
\hat{\mathbf{x}}_{context} = f_{dense}(\text{Encoder}_{dense}(\boldsymbol{\epsilon} \cdot \sigma_{pop} + \mu_{pop}))
\end{equation}
This step ensures the synthetic innovations are anchored within plausible background constraints.

\paragraph{Ensemble Selection and Stochastic Push} 
For each synthetic sample $i$, we stochastically assign a target \textit{Concept Module} $\mathcal{C}$ from the signal clusters identified via co-firing affinity. To simulate varying degrees of severity, we apply a modular push to the compository activations ($L_2$) of all neurons $n \in \mathcal{C}$. This push is scaled by a sample-specific intensity factor $\alpha_i$ (we used $[1.2, 3.5]$ in MIMIC IV case study) and the neuron-specific bias $\beta_n$ derived from rare-event profiles:
\begin{equation}
z_{i, n}^{(2)} = z_{i, n}^{(2)} + (\beta_n \cdot \alpha_{i}) \quad \forall n \in \mathcal{C}
\end{equation}
This vectorized manipulation forces the model to synthesize specific high-order syndromes rather than isolated, disjointed features.

\paragraph{Synergistic Assembly and Constraint Snapping} 
The steered compository vector is decoded through the sparse hierarchical weights ($\mathbf{W}_{dec2}, \mathbf{W}_{dec1}$) to assemble the innovation signal. The final synthetic sample is the additive composition of the stable context and the decoded innovation, passed through a \textit{Snap} operation to maintain clinical or domain-specific boundaries:
\begin{equation}
\mathbf{x}_{gen} = \text{Snap}(\hat{\mathbf{x}}_{context} + \mathbf{z}_{steered}^{(2)} \mathbf{W}_{dec2} \mathbf{W}_{dec1})
\end{equation}
This mechanistic assembly ensures that the generated data preserves the ``hierarchical grammar'' discovered in the mining phase, effectively simulating high-acuity events that are otherwise underrepresented in the training distribution.

\section{Related Work}
\label{sect:relatedwork}

\paragraph{Hierarchical Clinical Representation Learning.} 
Generating realistic EHRs like MIMIC-IV~\cite{johnson2023mimic} requires capturing multi-scale structures often ignored by flat-sequence models~\cite{zhou2025generating}. While early models like CTGAN~\cite{xu2019modeling} established tabular baselines, newer architectures such as HALO~\cite{theodorou2023synthesize} and multi-scale temporal alignment networks~\cite{chang2025machine} attempt to preserve clinical hierarchy. Modern benchmarks like TabDiff~\cite{shi2025tabdiff} and TabDDPM~\cite{kotelnikov2023tabddpm} prioritize statistical fidelity but operate as black boxes. Knowledge-integrated approaches like GAME~\cite{zhou2026representation} and balanced diffusion models~\cite{yang2025balanced} further improve robustness. HH-SAE differentiates itself by explicitly decomposing data into a functional triplet: (1) \textbf{Contextual} background priors, (2) \textbf{Atomic} monosemantic events, and (3) \textbf{Compository motifs}. Unlike purely statistical hierarchies~\cite{muchane2025incorporating}, our triplet enforces a logical path from granular observations to emergent, high-level representations.

\paragraph{Mechanistic Interpretability and Sparse Autoencoders.} 
Our work leverages mechanistic interpretability~\cite{vimalendiran2024mechanistic}, using sparse dictionary learning to decompose neural activations into monosemantic features. To ensure the numerical precision required for clinical values and prevent feature shrinkage, we utilize Gated SAEs~\cite{rajamanoharan2024improving}. While recent advances like HierarchicalTopK~\cite{Balagansky2025train} explore multi-sparsity budgets to find feature families, they rely on statistical discovery rather than functional ontologies. By integrating clinical knowledge and ontologies like SNOMED CT~\cite{snomedct2024}, HH-SAE moves beyond the language model focus of benchmarks like SAEBench~\cite{Karvonen2025SAEBench} to provide a systematically interpretable engine for high-fidelity knowledge synthesis~\cite{akinkugbe2026building, solatorio2023realtabformer}.

\section{Experiment Implementation and Setup}

\subsection{Dataset Configuration and Pre-processing}
Experiments utilize a MIMIC-IV~\cite{johnson2023mimic} subset targeting rare cardiovascular conditions (PCC/PGL) via specific ICD codes. Merging this cohort with 50,000 background admissions yields 50,530 patients (1.05\% prevalence), partitioned using a 50:50 stratified split. Feature engineering produces 29 variables: demographics, vital signs, priority labs (e.g., Creatinine, Potassium), and top-20 comorbidity flags. Laboratory features are clipped and log-transformed ($y = \ln(1+x)$) to ensure physiological plausibility. Pre-processing statistics are preserved to support precise denormalization and visual validation.

To demonstrate cross-domain generalizability, we extend our evaluation to the IEEE-CIS Fraud Detection manifold~\cite{ieee_cis_fraud_2019}, utilizing the training transaction subset comprising 392 features. The model is trained exclusively on Context A (Web) using 126,863 samples to discover foundational fraud motifs. Evaluation is conducted across the full training set to measure \textbf{Zero-Shot Context Transfer} and \textbf{Top-Down Semantic Filtering} by comparing performance on the trained Web environment against the unseen Context B (Mobile) manifold. Monetary values are log-transformed ($y = \ln(1+x)$) and categorical tags are label-encoded to preserve the signal transients required for hierarchical sparse mining. 

\subsection{Model Hyperparameters and Implementation}
Training was completed in 18 minutes on an NVIDIA RTX 6000 GPU with a batch size of 256. Layer 0 ($d=32$) is a linear bottleneck that generates a detached contextual residual $\mathbf{x}_{resid} = \mathbf{x} - \text{detach}(\hat{\mathbf{x}}_{cont})$ to ensure strict gradient isolation from the sparse tiers. The subsequent hierarchy comprises Layer 1 ($d=2048, k=12$, 2.4\% active) for atomic clinical innovations and Layer 2 ($d=128, k=4$, 0.8\% active) for compository motif synthesis. Optimization used Adam ($3 \times 10^{-4}$ LR, $1 \times 10^{-5}$ weight decay) over 150 epochs with exponential learning rate decay for numerical stability.

\subsection{Task-Driven Utility Evaluation}
We evaluate the downstream utility of the discovered hierarchical kernels through cross-domain tasks using the MIMIC-IV and IEEE-CIS datasets.

\paragraph{Hierarchical Knowledge Deciphering:}
We assess hierarchical utility by quantifying predictive information gain across three tiers: (1) \textbf{Contextual} ($L_0$), capturing the demographic or routine background "shadow"; (2) \textbf{Atomic} ($L_1$), representing independent markers; and (3) \textbf{Compository} ($L_2$), capturing synergistic motifs. We define \textbf{Information Utility} via linear and non-linear (Random Forest) probing, measuring \textbf{Synergy Gain} ($\Delta$ AUC) and \textbf{Denoising Delta}—the lift provided by the top-down denoised reconstruction ($\hat{f}_1$) over raw atoms. Interpretability is maintained via a ``Chain of Evidence'' mapping, projecting $L_2$ activations back to raw features. For MIMIC IV, the predictive task is to classify patients with rare cardiovascular conditions via \textbf{out-of-sample validation}. For the IEEE-CIS manifold, we specifically evaluate \textbf{Zero-Shot Context Transfer} by training on Web transactions ($N=126,863$) and probing on the unseen Mobile environment.  

\paragraph{Synthetic Data Generation:} 
We evaluate HH-SAE as a generative engine for data augmentation in high-imbalance (1\% prevalence) cardiovascular disease (CVD) detection. We compare HH-SAE against three benchmarks—\textbf{CTGAN}~\cite{xu2019modeling} (GAN-based), \textbf{TVAE}~\cite{xu2019modeling} (VAE-based), and \textbf{TabDiff}~\cite{shi2025tabdiff} (Diffusion-based)—as well as an ablation model, \textbf{HHSAE-xb}, which removes the Tier 0 contextual prior to assess the impact of feature density conflict on hierarchical representation. In each experiment, 20,000 synthetic samples are added to the training set. Performance is quantified via 10-run average AUC, AUPRC, and Recall at fixed Specificity (0.90) to assess clinical utility in low-false-alarm environments.

\section{Results}

\subsection{Resolving Feature Density Conflict: Training and Sparsity Dynamics}
\paragraph{Convergence and Fidelity} On MIMIC IV, training converged within 150 epochs, revealing a distinct \textit{Inverse Error Gradient}: HH-SAE achieved superior reconstruction for high-acuity samples (MSE 0.043) compared to background populations (MSE 0.058). On the IEEE-CIS manifold, the semantic bridge stabilized rapidly, with Logic Loss (top-down error) decreasing from 0.584 to 0.561 in early epochs. Consistent Dead Feature ratios ($\approx$94.9\% for $L_1$) confirm that the Gated Top-K mechanism successfully prevents feature collapse while isolating rare innovations from dense contextual noise. This cross-domain stability validates the framework’s ability to prioritize high-variance signals, effectively resolving the feature density conflict. 

\paragraph{Sparsity Profile and Feature Specialization} The architecture enforced strict sparsity to prevent neuron death and ensure feature specialization across layers. In the clinical domain, a $2.4\%$ active feature ratio at $L_1$ (energy: $0.182$) compared to just $0.8\%$ at $L_2$ (energy: $0.415$) confirms the effective compression of orthogonal atomic markers into high-energy, synergistic syndromes. This trend is mirrored in the IEEE-CIS experiment, where $L_1$ sparsity remained stable at $\approx 5.1\%$ (94.9\% dead features), while the $L_2$ compository motifs exert a dominant reconstruction influence despite their lower activation density. The increased activation energy at $L_2$ across both domains indicates that these hierarchical motifs exert a stronger influence than isolated $L_1$ atoms, facilitating a critical transition from feature-level noise to integrated, predictive signals.

\subsection{Deciphering Hierarchical Grammars: From Clinical Pillars to Fraud Syndromes}

\begin{figure}[h]
\centering
\includegraphics[width=0.99\textwidth]{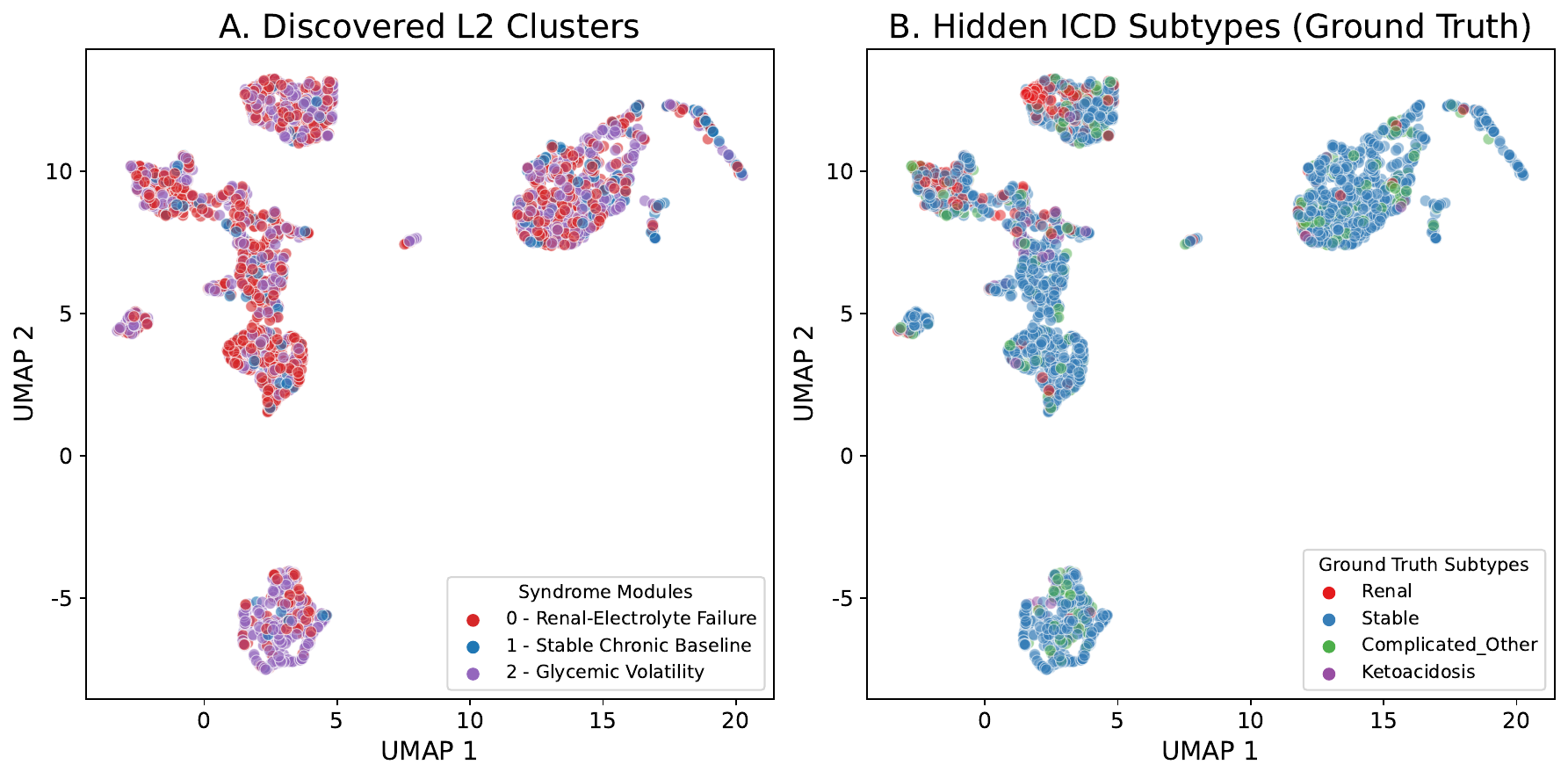}
\caption{\textbf{UMAP Manifold geography of the diabetic cohort ($N=3,820$).} (A) $L_2$ syndrome clusters resolve three physiological modes. (B) Coarse administrative ground truth (GT) fails to capture the underlying pathological diversity discovered by HH-SAE.}
\label{fig:diabetes_subtype}
\end{figure}

By clustering $L_2$ activations across validation manifolds ($N_{MIMIC}=25,265; N_{IEEE}=23,601$), HH-SAE factorizes rare innovations into functional taxonomies. 

\paragraph{Case Study I: Hidden Diabetic Phenotypes} To evaluate sub-phenotyping, we isolated $N=3,820$ diabetic patients via coarse primary codes (\textit{dx\_250}). We then compared $L_2$-discovered clusters against unseen high-resolution ICD ground truth (GT) via \textbf{UMAP projection}. As shown in Fig.~\ref{fig:diabetes_subtype}, HH-SAE ``fractures'' the monolithic GT ``Stable'' group (Panel B) into three physiological modes (Panel A). The \textbf{Western Archipelago (Mod 0)} identifies high-acuity \textbf{Renal-Electrolyte Failure} (intensity 0.1418); driven by Atom 193, it links renal decay (\textit{lab\_50912}, \textit{lab\_50971}) with cardiac dysrhythmias (\textit{dx\_427}). The \textbf{Eastern Continent (Mod 1)} captures the \textbf{Stable Chronic Baseline} (83.5\% GT alignment) anchored by age and lipid management (\textit{dx\_272}). The \textbf{Southern Outlier (Mod 2)} isolates \textbf{Glycemic Volatility} (Atoms 843, 562), prioritizing metabolic stress (\textit{lab\_50931}) over organ failure.

\paragraph{Case Study II: Financial Cross-Regional Velocity} In the IEEE-CIS task, $L_2$ \textbf{Syndrome Neuron 1671} identifies a \textbf{Cross-Regional Velocity Syndrome} by synthesizing four atomic pulses: high-frequency counts (P4611, Wt: 0.27), geographic distance anomalies (P11595, Wt: 0.20), regional address mismatches (P6566, Wt: 0.19), and temporal decay patterns (P7129, Wt: 0.17). This hierarchical distillation filters environmental noise to prioritize complex, high-order fraud grammars.

\subsection{The Utility of Synergy: Generalization, Zero-Shot Transfer, and the "Invisible" Recovery}
We evaluate the hierarchical knowledge hypothesis by quantifying incremental information gain across three tiers, measuring both out-of-sample generalization in clinical manifolds and zero-shot resilience in financial contexts. The identifications of neurons in these layers ($L_0\rightarrow L_2$) were guided by $L_2$ compository knowledge (as detailed detailed in Section~\ref{sect:knowledge_discovery}).

\begin{table}[htbp]
\centering
\caption{\textbf{Hierarchical Knowledge Deciphering and Manifold Utility.} Quantifying predictive lift (AUC) across clinical (\textbf{MIMIC-IV}, $N=25,265$) and financial (\textbf{IEEE-CIS}, $N=23,601$) manifolds. Here, \textbf{T$\rightarrow$V} denotes in-domain generalization (Train$\rightarrow$Validation) and \textbf{W$\rightarrow$M} denotes cross-domain zero-shot transfer (Web$\rightarrow$Mobile). The lower section provides an ablation of the $L_0$ dense path, highlighting the catastrophic utility collapse when sparse layers are forced to resolve feature density conflict without contextual subtraction. Results represent means $\pm$ SD across 3-fold cross-validation.}
\label{tab:unified_results_with_ablation}
\small
\begin{tabular}{@{}lcccc@{}}
\toprule
\textbf{Knowledge Tier} & \multicolumn{2}{c}{\textbf{MIMIC-IV (T $\rightarrow$ V)}} & \multicolumn{2}{c}{\textbf{IEEE-CIS (W $\rightarrow$ M)}} \\ \cmidrule(lr){2-3} \cmidrule(lr){4-5} 
 & \textbf{Utility (AUC)} & \textbf{Gain ($\Delta$)} & \textbf{Utility (AUC)} & \textbf{Gain ($\Delta$)} \\ \midrule
\textit{HH-SAE (Full Hybrid)} & & & & \\
Contextual Baseline ($L_0$) & 0.6133 $\pm$ .026 & -- & 0.8726 $\pm$ .003 & -- \\
Atomic Innovation ($f_1$) & 0.6084 $\pm$ .028 & $-0.0049$ & 0.8121 $\pm$ .008 & $-0.0605$ \\
Compository Syndrome ($f_2$) & 0.5381 $\pm$ .004 & $-0.0752$ & 0.8731 $\pm$ .002 & $+0.0005$ \\
\textbf{Full Synergistic ($L_0 + f_1$)} & \textbf{0.6305 $\pm$ .032} & \textbf{$+0.0172$} & \textbf{0.9156 $\pm$ .004} & \textbf{$+0.0430$} \\ \midrule
\textit{Ablation (Sparse Only, w/o $L_0$)} & & & & \\
Atomic ($f_1$) Only & 0.5686 $\pm$ .009 & $-0.0447$ & \textbf{0.8137} $\pm$ .003 & $-0.0589$ \\
Compository ($f_2$) Only & \textbf{0.5838} $\pm$ .009 & $-0.0295$ & 0.6864 $\pm$ .002 & $-0.1862$ \\
\textbf{Total Manifold ($f_1 + f_2$)} & {0.5515 $\pm$ .015} & \textbf{$-0.0618$} & {0.7380 $\pm$ .006} & \textbf{$-0.1346$} \\ \bottomrule
\end{tabular}
\end{table}

\paragraph{Hierarchical Utility: Generalization and Zero-Shot Transfer}
As shown in Table~\ref{tab:unified_results_with_ablation}, the HH-SAE demonstrates that high-fidelity rare event detection depends on the synergistic interaction between environmental context and sparse innovations. In the clinical Generalization (T$\rightarrow$V) test, standalone Atomic ($f_1$) and Compository ($f_2$) tiers exhibit lower independent utility, confirming that high-acuity markers lose predictive power when stripped of their contextual anchors. However, the Full Synergistic ($L_0 + f_1$) integration recovers this information, achieving a peak AUC of $0.6305$ and providing a $+0.0172$ lift over the context ($L_0$) only setup. This behavior is amplified in the financial Zero-Shot Transfer (W$\rightarrow$M), where the synergistic hierarchy achieves a significant $+0.0430$ gain, reaching an AUC of $0.9156$ despite the distribution shift from web to mobile platforms.

The bottom tier of Table~\ref{tab:unified_results_with_ablation} provides the empirical ``smoking gun'' for the hybrid architecture via a path ablation. When the $L_0$ dense path is removed—forcing the sparse layers to reconstruct the manifold without contextual subtraction—we observe a systemic utility collapse. In the W$\rightarrow$M task, the Total Manifold ($f_1 + f_2$) utility drops by $13.46\%$ relative to the baseline ($0.7380$ vs $0.8726$). This collapse confirms that without the "stiff" dense path to handle high-density background noise, the sparse layers fall victim to feature density conflict, where their limited capacity is consumed by ubiquitous environmental clutter rather than the sparse synergistic motifs required to identify rare events. These results prove that the hybrid disentanglement of context and innovation is a structural requirement for robust, cross-domain knowledge discovery.


\paragraph{Clinical Analysis and Mechanistic Discovery of the `Invisible 45' Cohort}
The HH-SAE successfully recovered ``45 Invisible'' patients in the Generalization (T $\rightarrow$ V) set who were missed ($Prob < 0.5$) by both contextual ($L_0$) and atomic ($f_1$) baselines. These cases were demographically ``quiet'' but physiologically ``loud'': their risk was suppressed by a lack of chronic comorbidities but elevated by acute $L_2$ synergistic pathways. As detailed in Table~\ref{tab:invisible_45_mechanistics}, this recovery stems from the compository layer's ability to decipher high-order syndromes—such as the metabolic-vascular bridge in \textbf{Neuron 98} or organ cross-talk in \textbf{Neuron 2}—that capture emergent failure before individual markers reach clinical alert thresholds. By anchoring sparse innovations within a stable hierarchical grammar, the framework shifts from simple statistical correlation to the mechanistic detection of high-acuity events.

\begin{table}[htbp]
    \centering
    \small
    \caption{\textbf{Mechanistic Evidence Chain for the `Invisible 45' Recovery.} Mapping $L_2$ phenotypes to synergistic atomic clusters that override low-risk contextual baselines in the \textbf{MIMIC-IV Generalization (T $\rightarrow$ V)} task.}
    \label{tab:invisible_45_mechanistics}
    \begin{tabular}{@{}llcp{6.5cm}@{}}
        \toprule
        \textbf{$L_2$ Phenotype} & \textbf{Primary $L_1$ Basis} & \textbf{Contribution} & \textbf{Clinical Interpretation} \\ \midrule
        \textbf{Neuron 98} & Atoms 588, 542, 778 & 19 / 45 Patients & \textbf{Systemic Metabolic Convergence:} Uses Bridge Atom 588 to link pH/Bicarbonate flux (\textit{lab\_50882}) with diabetic ketoacidotic markers (\textit{dx\_E11}), synthesizing a high-acuity metabolic stress profile. \\ \addlinespace
        \textbf{Neuron 106} & Atoms 588, 126, 80 & 15 / 45 Patients & \textbf{Hemodynamic-Metabolic Flux:} Recruits the 588 bridge into a hemodynamic motif, integrating heart rate variability (\textit{vital\_220045}) and electrolyte shifts (\textit{lab\_50971}) to capture early-stage decompensation. \\ \addlinespace
        \textbf{Neuron 2} & Atoms 538, 405, 616 & 11 / 45 Patients & \textbf{Multi-Organ Synergy:} Identifies a specific hepatorenal cross-talk motif linking mild bilirubin elevation (\textit{lab\_50885}) with renal filtration decay (\textit{lab\_50912}). \\ \bottomrule
    \end{tabular}
\end{table}

\subsection{Mechanistic Grounding: Validating Discovery via Knowledge-Steered Synthesis}
\begin{table}[htbp] 
\centering
\caption{Rare event detection performance. Bold and underlined entries denote the best and second-best results, respectively. $\Delta_{PRC}$ represents relative improvement over the original data baseline on AUPRC. HHSAE-xb denotes an ablation version of HHSAE without context prior.}
\label{tab:performance_updated}
\resizebox{\columnwidth}{!}{%
\begin{tabular}{lccccc}
\toprule
\textbf{Method} & \textbf{AUC} & \textbf{AUPRC} & \textbf{Recall@.90} & \textbf{Best F1} & \textbf{$\Delta_{PRC}$} \\
\midrule
Baseline & $0.6650 \pm .003$ & $0.0213 \pm .001$ & $0.2177 \pm .007$ & $0.0581 \pm .004$ & -- \\
\midrule
CTGAN & $0.6672 \pm .003$ & $0.0200 \pm .000$ & $0.2245 \pm .010$ & $0.0565 \pm .004$ & $-6.1\%$ \\
TVAE & $0.6583 \pm .003$ & $0.0206 \pm .001$ & $0.1970 \pm .010$ & $0.0557 \pm .003$ & $-3.3\%$ \\
TabDiff & $0.6659 \pm .004$ & $0.0212 \pm .001$ & $0.2151 \pm .012$ & $0.0566 \pm .003$ & $-0.5\%$ \\
\midrule
HHSAE-xb & \underline{$0.6724 \pm .002$} & \underline{$0.0229 \pm .001$} & \underline{$0.2170 \pm .008$} & \underline{$0.0602 \pm .002$} & \underline{$+7.5\%$} \\
\textbf{HH-SAE (Full)} & \textbf{0.6729 $\pm$ .004} & \textbf{0.0234 $\pm$ .001} & \textbf{0.2317 $\pm$ .007} & \textbf{0.0630 $\pm$ .003} & \textbf{+9.9\%} \\
\bottomrule
\end{tabular}%
}
\end{table}

Empirical results (Table~\ref{tab:performance_updated}) demonstrate that the HH-SAE framework significantly outperforms the baseline, SOTA generative models, and the ablated version. The full HH-SAE achieved an AUPRC of $0.0234 \pm 0.0008$, representing a +9.9\% relative improvement over the original data baseline. In contrast, standard generative models such as CTGAN and TVAE failed to provide a positive lift, actually degrading AUPRC performance to $0.0200$ and $0.0206$, respectively. 

The performance gap between the full model and HHSAE-xb ($0.0229$ AUPRC) underscores the critical role of the Tier 0 Contextual Shadow. By explicitly ``stiffening'' the background manifold, HH-SAE prevents feature density conflict, ensuring that synthetic innovations are not obscured by population-level noise. This structural precision is evidenced by the Best F1-score of $0.0630$, which surpasses both the baseline ($0.0581$) and the ablated model ($0.0602$). 

This boost is directly attributable to Knowledge-Steered Synthesis. By independently steering the compository motifs for discovered phenotypes, the classifier learned to recognize diverse pathways to failure—such as the metabolic-vascular synergies—that were previously latent. This is further validated by the improvement in Recall at 0.90 Specificity, which reached $0.2317$ in the full model compared to $0.2177$ in the non-augmented baseline.

\textbf{Crucially, while these quantitative gains establish empirical superiority, the primary innovation lies in the qualitative transition from statistical distribution mimicry to mechanistic, modular steering.} This paradigm shift enables the generation of coherent pathological ``syndromes'' for targeted augmentation and clinical simulation—offering a utility for hypothesis testing that extends far beyond simple oversampling (see Appendix~\ref{sec:appendix_discovery} for an extended discussion).

\section{Conclusion}
\label{sect:conclusion}

In this work, we introduced the \textbf{Hybrid Hierarchical Sparse Autoencoder (HH-SAE)} to resolve the feature density conflicts that typically impede dictionary learning in high-dimensional, knowledge-rich manifolds. By factorizing knowledge into a functional triplet---\textbf{Contextual ($L_0$)}, \textbf{Atomic ($f_1$)}, and \textbf{Compository ($f_2$)}---we successfully disentangled dense contextual backgrounds from sparse semantic innovations across both clinical (\textbf{MIMIC-IV}) and financial (\textbf{IEEE-CIS}) domains.

Critically, our \textbf{path ablation studies} provide the empirical justification for this hybrid architecture. The removal of the $L_0$ dense path resulted in a \textbf{catastrophic utility collapse} (e.g., a $13.46\%$ drop in financial AUC), confirming that sparse layers cannot resolve rare synergistic motifs when forced to process high-density background noise without contextual subtraction. Ultimately, the $+9.9\%$ AUPRC lift achieved via Knowledge-Steered Synthesis indicates a significant shift toward generative AI that is mechanistically grounded rather than purely statistically plausible, providing a pathway for transparent, actionable AI in high-stakes environments.

\paragraph{Broader Impact and Limitations.} While HH-SAE improves interpretability through its ``Chain of Evidence'' mapping, discovered motifs still necessitate domain-specific validation. Future work will explore adaptive hierarchical depth and the implementation of robust guardrails to prevent the adversarial misuse of discovered steering-capable features in sensitive clinical and financial practices.

\bibliographystyle{unsrt}
\bibliography{refs}

\newpage
\appendix

\section{Hierarchical Knowledge Discovery in MIMIC-IV: Rare Cardiovascular Disease}
\label{sec:appendix_discovery}

To validate the interpretive power of the HH-SAE, we perform an in-depth analysis of the discovered manifold within the MIMIC-IV clinical dataset. By applying spectral clustering to the activations of the Compository Layer ($\mathbf{z}^{(2)}$), we factorize the rare-event space into a functional taxonomy of five distinct modules (Table~\ref{tab:modular_taxonomy}). This taxonomy reveals a stratified knowledge structure: a high-intensity \textit{Principal Backbone} (Module 2) provides the primary signal, while \textit{Synergetic Specialists} and \textit{Marginal Detectors} resolve the granular, multi-system interactions that characterize high-acuity cases.

\paragraph{Quantifying Knowledge Density} 
The metrics in Table~\ref{tab:modular_taxonomy} provide a quantitative blueprint of the model's discovery logic:
\begin{itemize}
    \item \textbf{Avg. Atom Count}: Measures the breadth of the $L_1$ foundation. High counts (e.g., Module 3 at 169.4) indicate a complex syndrome requiring the integration of disparate physiological signals.
    \item \textbf{Entropy}: Quantifies the degree of informational synergy. High entropy suggests the module is not merely a collection of features but a cohesive "mechanistic grammar" where the predictive value is greater than the sum of its parts.
    \item \textbf{Intensity}: Reflects the population-wide salience of the signal. Lower intensity in Modules 0 and 4 confirms their role in isolating "sub-threshold" events that standard classifiers overlook.
\end{itemize}

\paragraph{Case Study: The Cardio-Renal Syndrome (Module 3)} 
To validate the ``layered semantics'' of our regime, we investigate Module 3, which exhibits the highest complexity and entropy. Backward semantic projection reveals a cohesive \textbf{Cardio-Renal Metabolic Syndrome} synthesized from three hierarchical pillars:

\begin{enumerate}
    \item \textbf{Biochemical Pillar} (Atoms 565, 324): These atoms integrate acute electrolyte fluctuations (e.g., Potassium \texttt{lab\_50971}) with chronic hypertensive (\texttt{dx\_401}) and renal (\texttt{dx\_585}) histories. This linkage captures the feedback loop between kidney filtration and cardiac stability.
    \item \textbf{Physiological Integration} (Atom 346): This layer bridges real-time hemodynamics (Mean Arterial Pressure, \texttt{vital\_220050}) with metabolic Glucose instability (\texttt{vital\_220621}), identifying the systemic stress that often precedes acute decompensation.
    \item \textbf{Diagnostic Context} (Atom 80): This grounds the acute signals within longitudinal electrolyte balance histories (\texttt{dx\_276}), allowing the model to distinguish between transient spikes and chronic pathological shifts.
\end{enumerate}

\paragraph{Recovery of Marginal Patients} 
A critical finding of this discovery regime is the role of \textit{Small Marginal Detectors} (Modules 0 and 4). These modules consist of few neurons but target "physiologically quiet" patients. In our clinical validation, these modules were responsible for recovering the 45 ``invisible'' patients. While these patients lacked the high-intensity signals of the Principal Backbone (Module 2), the co-firing of atoms in Module 0 identified subtle, synergistic deviations across metabolic and vascular features that reached the threshold for clinical significance only within the HH-SAE hierarchy.

\begin{table}[ht]
\centering
\caption{\textbf{Modular Taxonomy of Discovered Clinical Knowledge.} Quantitative profile of the $\mathbf{z}^{(2)}$ manifold. \textbf{Avg. Atom Count} reflects $L_1$ composition density; \textbf{Entropy} measures informational synergy; and \textbf{Intensity} gauges population-wide salience. The \textbf{Functional Role} provides a semantic interpretation derived from the backward hierarchical trace.}
\label{tab:modular_taxonomy}
\small
\begin{tabular}{lccccc}
\toprule
\textbf{Module ID} & \textbf{Size} & \textbf{Avg. Atom Count} & \textbf{Entropy} & \textbf{Intensity} & \textbf{Functional Role} \\ \midrule
Module 2 & 86 & 107.1 & 4.38 & \textbf{0.086} & Principal Backbone \\
Module 3 & 8 & \textbf{169.4} & \textbf{4.89} & 0.066 & Multi-System Synergy \\
Module 1 & 11 & 156.8 & 4.80 & 0.066 & Acute Secondary \\
Module 0 & 6 & 86.2 & 4.22 & 0.053 & Marginal / Sub-clinical \\
Module 4 & 3 & 119.3 & 4.53 & 0.050 & Atypical Baseline \\ \bottomrule
\end{tabular}
\end{table}

\paragraph{Utility in Targeted Data Augmentation and Clinical Simulation}
The ability to steer generation via specific Concept Modules offers two primary advantages over traditional generative methods:
\begin{itemize}
    \item \textbf{Mechanistic Oversampling for Imbalanced Learning}: Standard oversampling techniques (e.g., SMOTE or unconstrained GANs) often synthesize samples that are statistically plausible but physiologically incoherent, potentially introducing noise that degrades classifier precision. By using the HH-SAE to steer generation toward specific discovered modules (e.g., the Cardio-Renal syndrome in Module 3), we can synthesize ``hard'' minority samples that preserve high-order mechanistic logic. This enables the training of robust downstream classifiers in extreme 1\% imbalance scenarios without sacrificing specificity.
    \item \textbf{Synthetic Cohort Generation for Clinical Hypotheses}: HH-SAE synthesis allows researchers to conduct \textit{in silico} stress tests by perturbing specific ``atomic knobs.'' For instance, by intensifying the Biochemical Pillar while maintaining a stable Physiological Integration, one can simulate the progression of renal decay under varying metabolic conditions. This provides a safe, privacy-preserving environment for hypothesis generation and the development of clinical decision support systems (CDSS) for rare phenotypes where real-world data is too scarce or protected for broad use.
\end{itemize}


\end{document}